\documentclass{article}
\usepackage{ijcai20}

\usepackage{times}

\usepackage{soul}
\usepackage{url}
\usepackage[hidelinks]{hyperref}
\usepackage[utf8]{inputenc}
\usepackage[small]{caption}
\usepackage{graphicx}
\usepackage{amsmath}
\usepackage{booktabs}
\usepackage{amssymb}
\usepackage{subfigure}
\title{BARNet: Bilinear Attention Network with Adaptive Receptive Fields for Surgical Instrument Segmentation}

\author{
Zhen-Liang Ni$^{1,2}$\and
Gui-Bin Bian$^{1,2}$\footnote{Contact Author}\and
Guan-An Wang$^{1,2}$\and
Xiao-Hu Zhou$^1$\and\\
Zeng-Guang Hou$^{1,2,3}$\and
Xiao-Liang Xie$^1$\and
Zhen Li$^1$\And
Yu-Han Wang$^1$
\\
\affiliations
$^1$Institute of Automation, Chinese Academy of Sciences, Beijing, China\\
$^2$The School of Artificial Intelligence, University of Chinese Academy of Sciences, Beijing, China\\
$^3$CAS Center for Excellence in Brain Science and Intelligence Technology, Beijing, China\\
\emails
\{nizhenliang2017\}@ia.ac.cn
}

\begin{document}

\maketitle

\begin{abstract}
Surgical instrument segmentation is extremely important for computer-assisted surgery. Different from common object segmentation, it is more challenging due to the large illumination and scale variation caused by the special surgical scenes. In this paper, we propose a novel bilinear attention network with adaptive receptive field to solve these two challenges. For the illumination variation, the bilinear attention module can capture second-order statistics to encode global contexts and semantic dependencies between local pixels. With them, semantic features in challenging areas can be inferred from their neighbors and the distinction of various semantics can be boosted. For the scale variation, our adaptive receptive field module aggregates multi-scale features and automatically fuses them with different weights. Specifically, it encodes the semantic relationship between channels to emphasize feature maps with appropriate scales, changing the receptive field of subsequent convolutions. The proposed network achieves the best performance 97.47$\%$ mean IOU on Cata7 and comes first place on EndoVis 2017 by 10.10$\%$ IOU overtaking second-ranking method.

\end{abstract}


\section{Introduction}
In recent years, there has been significant progress in minimally invasive robotic surgery and computer-assisted microsurgery. Semantic segmentation of surgical instrument plays a crucial role in assisted surgery. It can accurately locate the surgical instrument and estimate its pose, which is essential for surgical robot control~\cite{endovis2017}. Furthermore, the mask generated by semantic segmentation offers numerous solutions to assist surgery, such as real-time surgical reminder, objective assessment of surgical skills, surgical report generation and surgical workflow optimization~\cite{Sarikaya}. These applications can improve the safety of surgery and reduce the workload of doctors.

\begin{figure}[tbp]
  \centering
  \includegraphics[width=0.98\columnwidth]{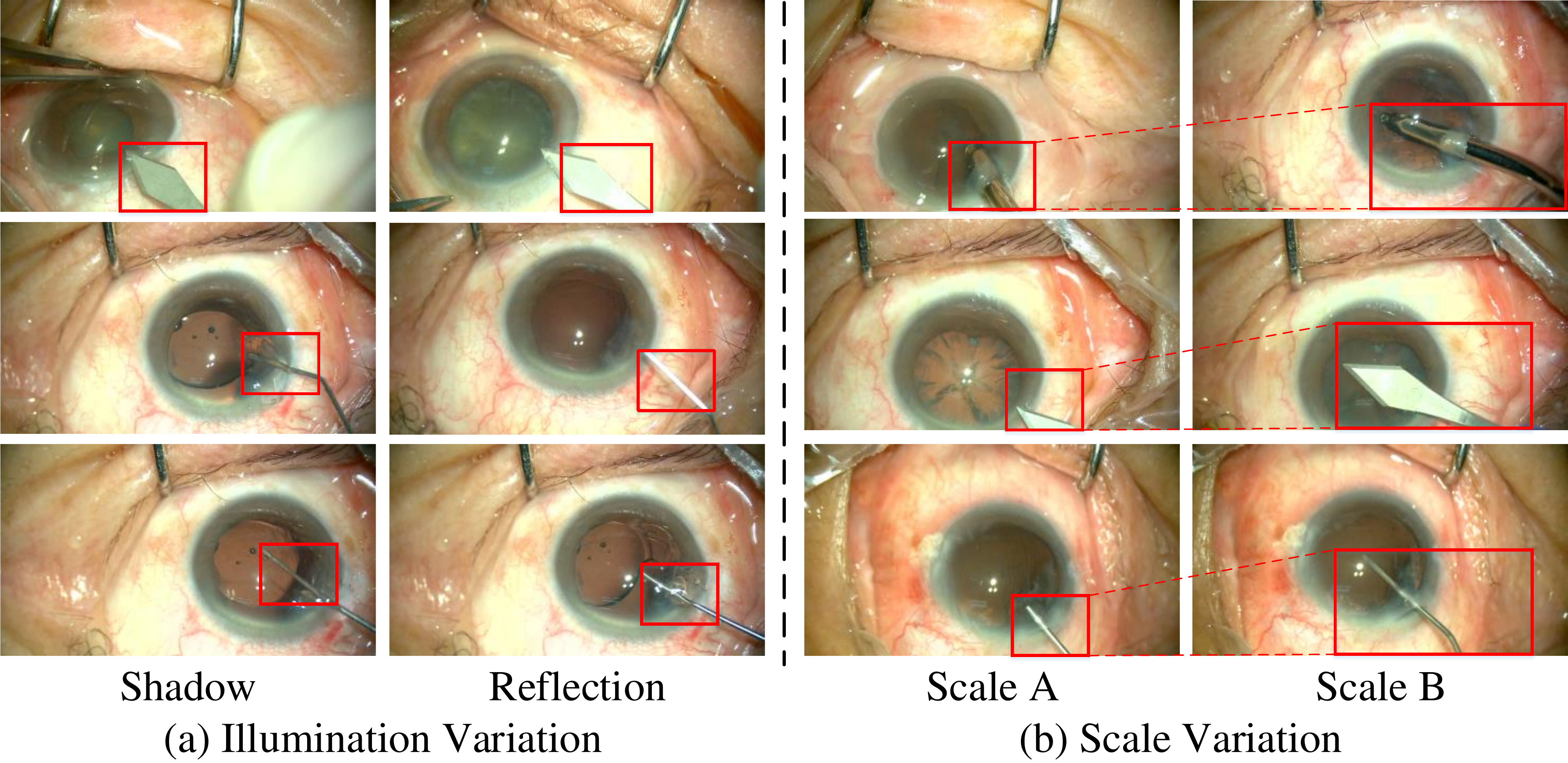}
  \caption{Difficulties in semantic segmentation for surgical instruments. Illumination variation changes the color and texture of instruments. Scale variation changes the size and shape of instruments.}
  \label{show}
\end{figure}

Compared with common object segmentation, the complex surgical scenes make accurate instrument segmentation more challenging. The first difficulty is the large illumination variation caused by the different light angles and occlusions. As shown in Figure~\ref{show} (a), surgical instruments tend to be whiter under specular reflection while the shadow makes instruments and background black. These problems seriously affect the visual representation of surgical instruments such as color and texture, impeding stably identifying the instrument. The second difficulty is the large scale variation caused by continuous movement and view changes. It leads to different shapes and scales of the same instrument. For example, the incision knife is in the shape of a triangle when the scale is small and in the shape of a polygon when the scale is large in Figure~\ref{show} (b). This issue makes the instrument segmentation more challenging.

Recently, a series of methods have been proposed for the semantic segmentation of surgical instruments. RAUNet~\cite{raunet} designed an attention module to fuse multi-level feature maps and emphasize the target region. A hybrid CNN-RNN method~\cite{attia} introduced Recurrent Neural Network to capture global contexts and expand the receptive field. MF-TAPNet~\cite{mf-tapnet} adopted optical flow as temporal prior to provide a reliable indication of the instrument location and shape for accurate segmentation. ToolNet-C combined with the kinematic pose information to get the accurate silhouette mask~\cite{qin}. However, most of those work focuses on expanding the receptive field and capturing shape prior while fail to address the scale variation and illumination variation issues.

To address the issues mentioned above, we reconsider the features affected by them. Illumination variation affects the color and texture appearance, making identifying instruments harder. Considering that a surgical instrument is spatially continuous, we can infer the target region according to its neighbor pixels based on semantic dependency and global context. To this end, a bilinear attention module (BAM) is proposed, which is based on bilinear pooling to model semantic dependencies and aggregate global contexts. Bilinear pooling can capture second-order feature statistics to encode complex semantic dependencies, helping to improve feature representations. Furthermore, attention features generated by bilinear pooling are adaptively distributed to each location, making every pixel feel global contexts. In this way, semantic features in reflective or shaded areas can be inferred based on semantic dependencies and global contexts, dealing with the illumination variation.

Besides, the scale variation changes the shape and size of surgical instruments.
Thus, we propose an adaptive receptive field module (ARF) to select and merge receptive fields of different scales adaptively. By doing so, we can cover various scales and make predictions more reliable.
Specifically, ARF includes two branches. The former learns semantic relationships among channels, and the latter aggregates multi-scale features.
Channel-wise semantic relationships are applied to select feature maps with appropriate sizes. Since kernels with the same size have different receptive fields on feature maps with various sizes, this module can select appropriate receptive fields for instruments at different scales by selecting specific feature maps, adapting to the scale variation. Moreover, dense connections across scales are introduced to propagate multi-scale features, which can cover a larger scale range.

Based on the above analysis, the bilinear attention network with adaptive receptive field, named BARNet, is proposed.
The contributions of this work are as follows:
\begin{itemize}
\item We propose the bilinear attention module to model semantic dependencies and aggregate global contexts for inferring the semantic features in the challenging region.

\item We design the adaptive receptive field module to select the appropriate receptive field adaptively, adapting to the scale variation of instruments.

\item The proposed network achieves state-of-the-art performance 97.47\% mIOU on Cata7 and comes first place on EndoVis 2017 by 10.10$\%$ IOU overtaking second place.
\end{itemize}
\section{Related Work}


\subsection{Attention}
Attention mechanisms are widely used in semantic segmentation tasks. Some attention models extract attention features based on first-order operations such as global average pooling and convolution pooling. For instance, SENet~\cite{senet} applied global average pooling to capture global contexts and model semantic dependencies between channels. AGRNet~\cite{zhang} utilized convolution pooling to generate attention features. Besides, some works applied second-order models to encode complex semantic dependencies~\cite{a2net}. For example, A2Net~\cite{a2net} was based on bilinear pooling to capture second-order statistics and model semantic dependencies. The bilinear attention networks~\cite{ban} learned bilinear attention distributions, on the top of the low-rank bilinear pooling technique. These work suggested that bilinear models can capture abundant semantic relationships to improve feature representation. Different from the above methods, we design a novel encoder-decoder architecture to aggregate and distribute attention features, which can achieve better results.


\subsection{Adaptive Receptive Field and Pyramid Features}
Pyramid features play a critical role in the segmentation of multi-scale targets. A range of methods improved feature representation by aggregating multi-scale features. For example, PSPNet~\cite{pspnet} adopted pyramid pooling to extract multi-scale features. DeepLabV2~\cite{deeplabv2} made use of dilated convolutions with different dilation rates to achieve multi-scale features. Feature pyramid network~\cite{fpnet} laterally propagated multi-scale features for building feature maps with rich semantic information at all scales.
However, these methods only concatenate multi-scale features together while they fail to select appropriate scales for a specific task. SKNet utilized different sized kernels to generate feature maps with different receptive fields and filter them~\cite{sknet}, which provided a solution for the adaptive selection of receptive fields. Different from the above methods, we directly select multi-scale features instead of different size kernels and do not need to generate new features, reducing calculation costs. Besides, multi-scale features are generated by dense connections across scales, which can cover more scale ranges and boost the reuse of multi-scale features~\cite{denseaspp}.

\section{BARNet}
\subsection{Overview}
\begin{figure}[tbp]
\centering
\includegraphics[width=0.96\columnwidth]{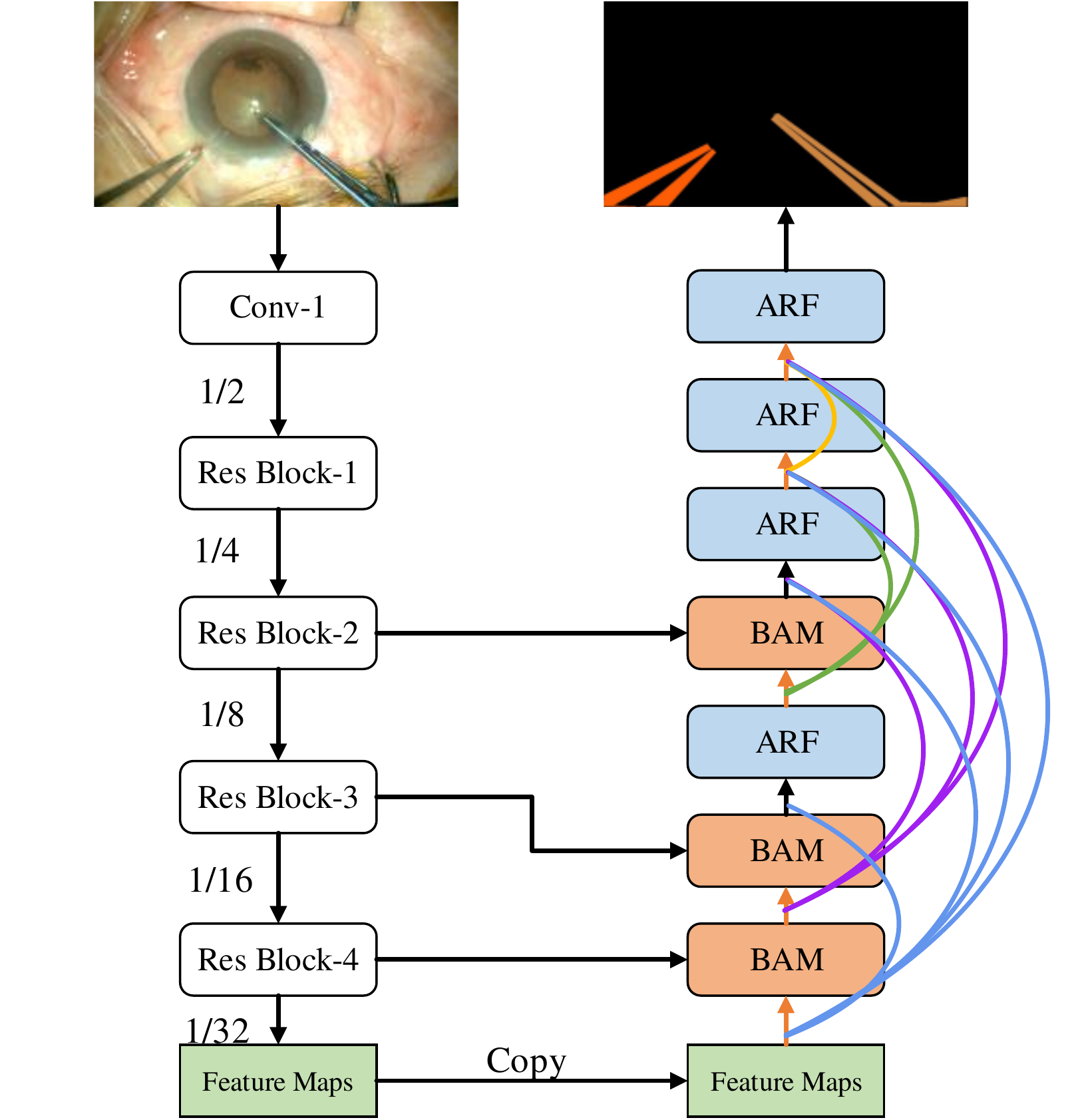}
\caption{The architecture of BARNet. It contains two critical modules, the bilinear attention module and the adaptive receptive field module. Dense connections across scales are introduced to propagate multi-scale features.}
\label{overall}
\end{figure}

Illumination variation leads to the change in color and texture of instruments. Scale variation varies the size and shape of instruments. To address these issues, a bilinear attention network with adaptive receptive field is proposed to capture semantic dependencies and global contexts for inferring semantic features in challenging areas and select receptive field adaptively for adapting to the scale variation.
The overall network architecture is shown in Figure~\ref{overall}. The bilinear attention module models semantic dependencies in small scale feature maps and generates attention features. The adaptive receptive field module aggregates multi-scale features and selects appropriate receptive field for instruments at different scales. Besides, cross-scale dense connections are introduced to propagate multi-scale features, which can also improve information flow and boost the reuse of features.
Since the input of adaptive receptive field module should be multi-scale features, we do not use this model when the feature map is small. Residual Network pre-trained on the ImageNet is adopted as the backbone network.

\subsection{Bilinear Attention Module}
\begin{figure}[tbp]
\centering
\includegraphics[width=\columnwidth]{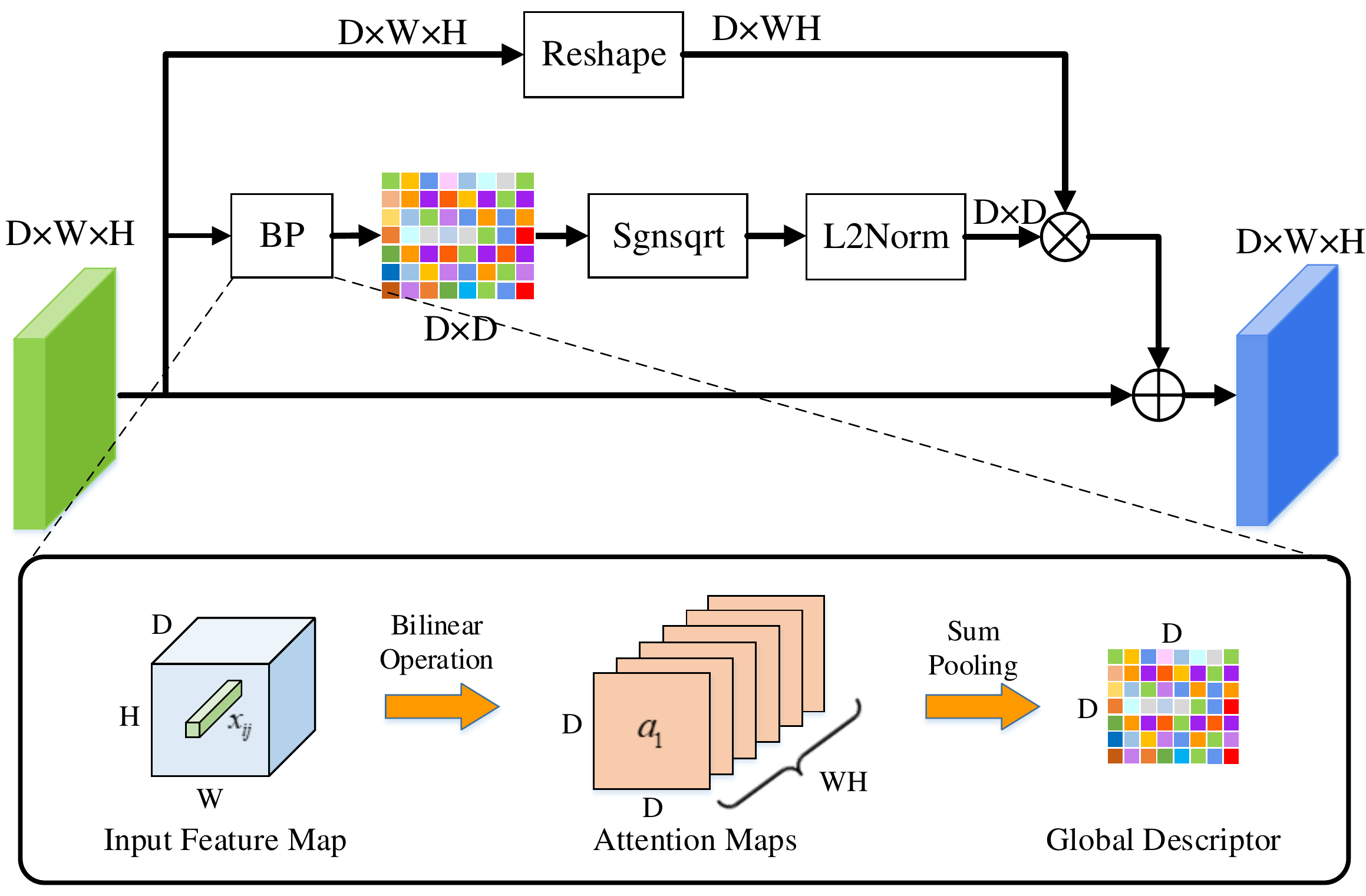}
\caption{The architecture of bilinear attention module. $\otimes$ denotes matrix multiplication. $\oplus$ denotes matrix addition.}
\label{attention}
\end{figure}

Illumination variation leads to the change in color and texture of surgical instruments. The network cannot utilize these features to identify surgical instruments, making it difficult to segment them. To solve this issue, the bilinear attention model is proposed to model semantic dependencies and captures global contexts. It is based on bilinear pooling to capture second-order statistics~\cite{bcnn}, which helps to boost the distinction between different semantic features. Thus, the bilinear attention module can encode more complex dependencies. Besides, a decoder is designed to distribute global contexts to each location adaptively.
The bilinear attention module is shown in Figure~\ref{attention}, including three parts: encoding, normalization, and decoding.
\begin{equation}
{Z} = {F_{decode}}\left( {{F_{norm}}({F_{bp}}(X))} \right)
\end{equation}

The first step is to model semantic dependencies and capture global contexts. Bilinear pooling is utilized to achieve this goal. It calculates the outer product of the feature vector pair $\left( {{x_{ij}},{y_{ij}}} \right)$ to capture second-order statistics and generate attention map ${a_{ij}}$. Each attention map represents the features of a pixel. These attention maps are concatenated together to encode spatial semantic dependencies.
Then, sum pooling is performed to generate the global descriptor $A \in {R^{D \times D}}$, as illustrated in Eq.(2). Semantic features in all locations are encoded into each element of the global descriptor, making each element feel global information. In this way, the bilinear attention module encodes semantic dependencies and aggregates global contexts.
\begin{equation}
A={F_{bp}}(X,Y) = {X{Y^T}} = {\sum\limits_{i = 1}^W {\sum\limits_{j = 1}^{\rm{H}} {{x_{ij}}{y_{ij}}^T} } }
\end{equation}
where $X\in {R^{D \times W \times H}}$ and $Y \in {R^{D \times W \times H}}$. In this work, we set $X=Y$.

Then, the global attention map $A \in {R^{D \times D}}$ is normalized to further improve its feature representation. The element-wise signed square-root
and ${\ell _2}$ normalization
are performed to normalize it, which can improve performance in practice. Also, these
operations are piecewise differentiable, which can be applied to end-to-end training~\cite{bcnn}.
\begin{equation}
A' = {F_{norm}}(A{\rm{) = }}{{{sign(A)\sqrt {\left| A \right|} } \mathord{\left/
 {\vphantom {{sign(A)\sqrt {\left| A \right|} } {\left\| {sign(A)\sqrt {\left| A \right|} } \right\|}}} \right.
 \kern-\nulldelimiterspace} {\left\| {sign(A)\sqrt {\left| A \right|} } \right\|}}_2}
\end{equation}

The last step is to distribute attention features to each pixel of the input feature map, making semantic features of each pixel calibrated by global information. The input feature map $X$ is reshaped into $\overline X  \in {R^{D \times WH}}$. By applying matrix multiplication, attention features are distributed to each location of the input feature map. The attention feature map $Z \in {R^{D \times W\times H}}$ is achieved, as shown in Eq.(4).
\begin{equation}
Z = {F_{decode}}(A,X) = A' \times \bar X + X
\end{equation}

The bilinear attention module allows each pixel of feature maps to feel global contexts. Semantic features in reflective or shaded areas can be inferred based on global contexts and semantic dependencies, dealing with the illumination variation issue. Besides, it encodes semantic dependencies in the form of second-order statistics, which boosts the distinction between various semantics and improves feature representation. Furthermore, this module only performs matrix operations and does not add any parameters, which can be easily inserted into other networks.

\subsection{Adaptive Receptive Field Module }
Since the surgical instrument is continually moving during the surgery, its shape and scale are constantly changing. Adaptive receptive fields can help the network adapt to scale variation and learn more detailed features. Thus, the adaptive receptive field module is proposed to aggregate multi-scale features and select the appropriate receptive field for instruments at different scales. The global average pooling is introduced to model the semantic relationship between channels and generate the weight vector. The weight vector can highlight feature maps which have an appropriate scale. Since kernels with the same size have different receptive fields on feature maps with different scales, by selecting feature maps in a specific size, the receptive field of subsequent convolutions can be determined. In this way, the adaptive receptive field module can select the receptive field adaptively according to the semantic relationship between channels.

\begin{figure}[tbp]
\centering
\includegraphics[width=0.49\textwidth]{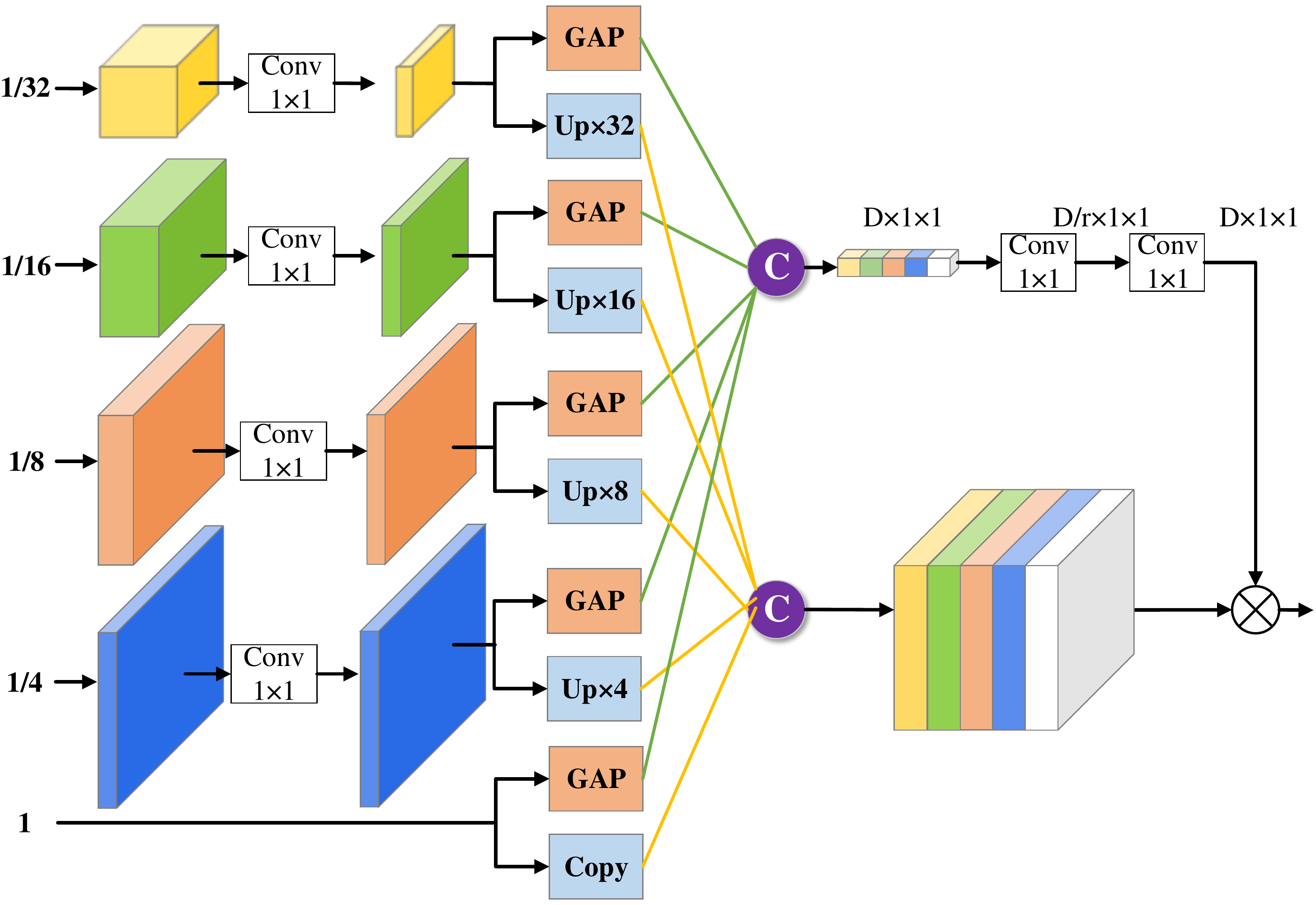}
\caption{The architecture of  adaptive receptive field module. $\otimes$ denotes broadcast Hadamard product.}
\label{ARF}
\end{figure}

The adaptive receptive field module is illustrated in Figure~\ref{ARF}. Take an input feature map with five scales as an example. First, 1$\times$1 convolution is performed on low-scale feature maps to adjust the channel dimension to $N$, which contributes to the aggregation of multi-scale features and reducing computational costs. $N$ can be selected according to the complexity of the network. The maximum scale feature map does not compress the channels to preserve semantic features as much as possible. In this paper, we set $N$ to 8. Then, multi-scale features are fed into two branches, one of which models the semantic relationship between channels and another one aggregates multi-scale features.

Specifically, in the first branch, multi-scale features are fed into global average pooling to generate vectors that encode semantic dependencies~\cite{senet}. These vectors are concatenated together and go through two convolution layers to further extract the semantic relationship. In this way, we obtain the weight vector which represents the degree of semantic responses for different feature maps. In the second branch, multi-scale features are aggregated by upsampling and concatenation, generating the pyramid feature. The pyramid feature is multiplied by the weight vector to select the feature map with a larger response. In this way, it can adjust the receptive field of subsequent convolutions adaptively.
\begin{equation}
{P_S} = S \otimes P
\end{equation}
where $\otimes$ refers to broadcast Hadamard product, $P$ represents the pyramid feature and $S$ denotes the attention vector.
\begin{equation}
P = H\left( {\left[ {{f_{{2^k}}}({x_0}),{f_{{2^{k - 1}}}}({x_1}), \ldots ,{f_{{2^1}}}({x_{k - 1}})} \right]} \right)
\end{equation}
where $x_k$ represents the $k$-th low-scale feature map. $f_{{2^k}}$ denotes the 1$\times$1 convolution and the ${2^k} \times$ upsampling. $H$ refers to concatenation.

\subsection{Loss Function}
To address the class imbalance issue, we use a hybrid loss consisting of cross-entropy and Dice~\cite{raunet}. Cross-entropy is often used for classification tasks. However, it is easily affected by the class imbalance issue, which leads to poor training results. Thus, Dice loss is introduced to address this problem. Dice evaluates the similarity between the ground truth and the output, which has no relation to the ratio of foreground pixels to background pixels. It is not affected by the class imbalance issue. This hybrid loss merges cross-entropy and Dice in a new way to effectively utilize their excellent characteristics, which is shown in Eq.(7).
\begin{equation}
Loss = (1-\alpha)H - \alpha \ln (D), \alpha  \in [0,1]
\end{equation}
where $H$ refers to cross-entropy and $D$ denotes Dice loss. $\alpha$ is a weight used to balance cross-entropy and Dice loss. It is set to 0.2 for best training results.

\section{Experiments}

\subsection{Dataset}
A cataract surgical instrument dataset, called Cata7, is used to evaluate our network. This dataset contains seven cataract surgery videos. To reduce redundancy, each video is downsampled from 30 fps to 1 fps. The resolution of the image is 1920$\times$1080 pixels. The entire dataset contains 2500 images, 1800 of which are used for training and the others are used for testing. There are 10 cataract surgical instruments in this dataset.

EndoVis 2017 dataset is from 2017 MICCAI Endovis Robotic Instrument Segmentation Challenge. This dataset is based on endoscopic surgery. All videos are acquired by a Vinci Xi robot. It contains 3000 images with a resolution of 1280$\times$1024, which contains 1800 images for training and 1200 images for the test. There are 7 types of surgical instruments in EndoVis 2017.

\subsection{Implementation Details}
All experiments are implemented on two Nvidia Titan X. The Residual Network pre-trained by ImageNet is used as the encoder. Adam with batch size 8 is used to train our network. The learning rate is dynamically adjusted during training. The initial learning rate on Cata7 dataset is $8 \times {10^{ - 6}}$ and the initial learning rate on EndoVis 2017 dataset is $5 \times {10^{ - 5}}$. For every 30 iterations, the learning rate is multiplied by 0.8. Due to limited computing resources, each image in the Cata7 is resized to 960$\times$544 pixels, and images in EndoVis 2017 are resized to 640$\times$512.
Data augmentation is performed to increase sample diversity. The selected samples are randomly rotated, shifted and flipped. 800 images are obtained by data augmentation.
To objectively evaluate our model, Dice and Intersection-over-Union(IoU) are selected as the evaluation metric.

\begin{figure*}[tbp]
\centering
\includegraphics[width=\textwidth]{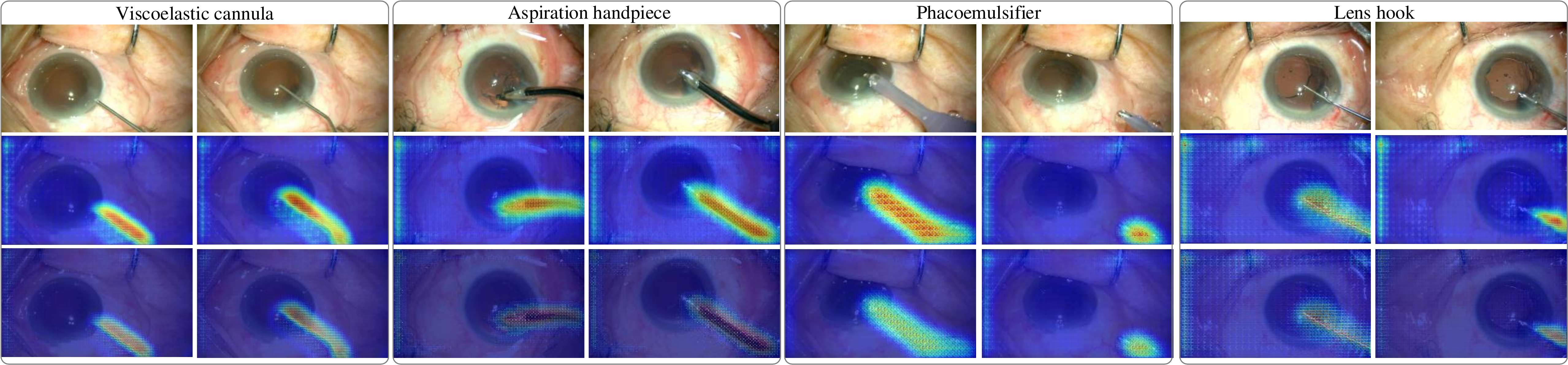}
\caption{Visualization of attention feature maps. From top to bottom: original image, attention feature maps generated by bilinear attention module, input feature map of bilinear attention module. Compared with input feature maps, attention features highlights the area where the instrument is located, indicating that the bilinear attention module can effectively capture the semantic features of the target region.}
\label{hotmap}
\end{figure*}
\subsection{Ablation Study Based on Cata7}
\subsubsection{Ablation Study for Bilinear Attention Module}
Bilinear attention module (BAM) is introduced to capture the second-order statistics and model long-range semantic dependencies. To verify its performance, some experiments are performed, as shown in Table~\ref{ablation}.
\begin{table}[htbp]
  \centering
    \resizebox{\columnwidth}{!}{
    \begin{tabular}{c|c|c|c|c|c}
    \hline
    \hline
    \textbf{Method} & \textbf{ARF} & \textbf{BAM} & \textbf{mDice(\%)} & \textbf{mIOU(\%)} &\textbf{Param.} \\
    \hline
    Basic &       &       &    95.12   &  91.31 & 21.80M  \\
    Basic &       & \checkmark    &  97.81     &   95.97 &  21.80M  \\
    Basic & \checkmark     &       &  98.06   &  96.28  &  21.90M\\
    Basic & \checkmark     & \checkmark     &   98.68    &  97.47  & 21.90M\\
    \hline
    \hline
    \end{tabular}
    }%
  \caption{Ablation experiments for Bilinear Attention Module and Adaptive Receptive Field Module on Cata7.}
  \label{ablation}%
\end{table}%

BARNet without BAM and ARF is used as the basic network. Compared with the basic network, the network using BAM has achieved an increase of 4.66$\%$ mean IOU and 2.69$\%$ mean Dice. When using ARF, employing BAM brings a 1.19$\%$ increase on mean IOU and 0.62$\%$ increase on mean Dice. It should be noted that BAM does not add any parameters, as shown in Table~\ref{ablation}. These experiments demonstrate that BAM can significantly improve network performance without adding any parameters.

\begin{figure}[tbp]
\centering
\includegraphics[width=0.98\columnwidth]{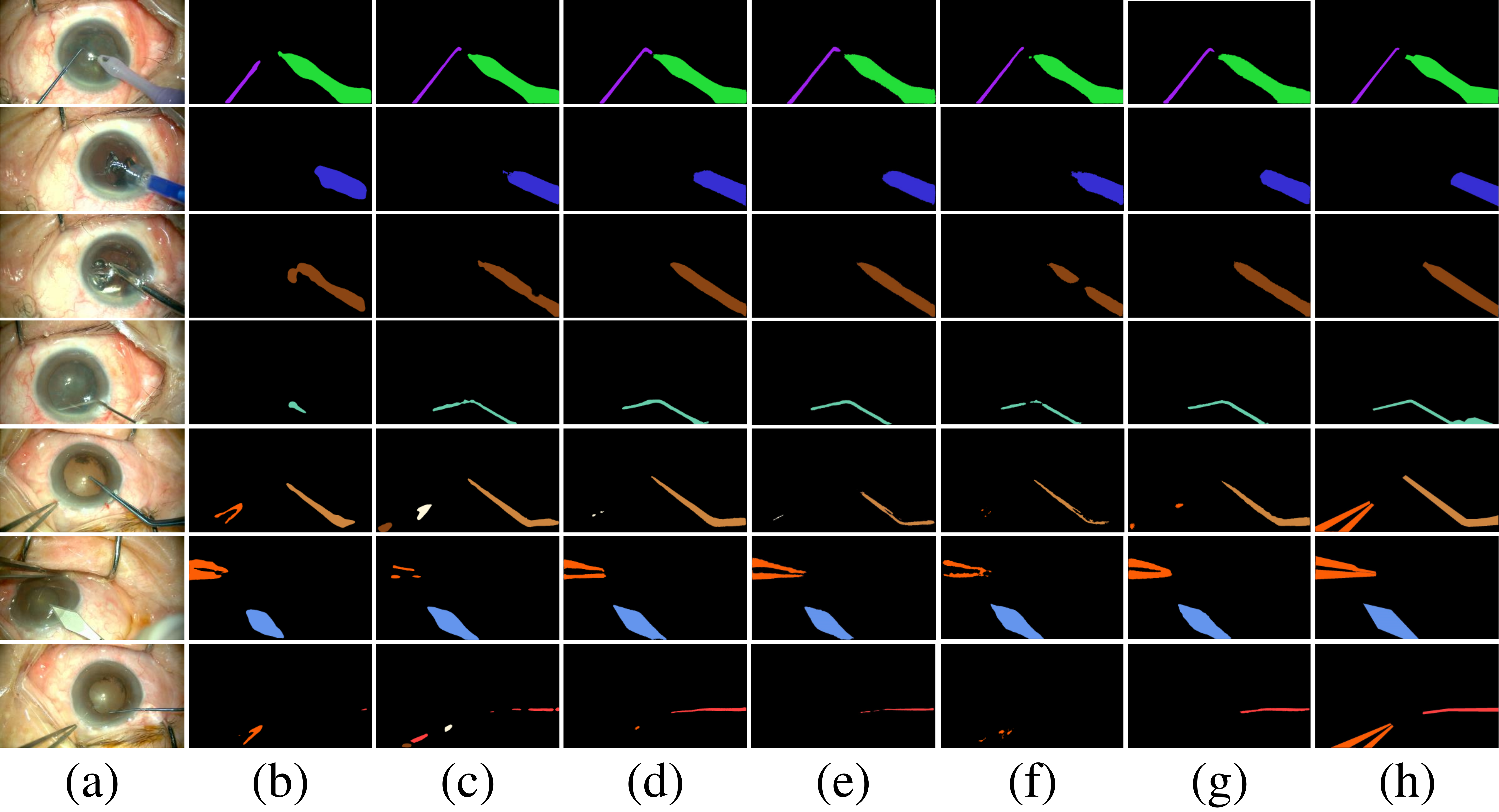}
\caption{Visualization of segmentation results for different methods. (a) Image; (b) U-Net; (c) TernausNet; (d) LinkNet; (e) Without BAM; (f) Without ARF; (g) BARNet (Ours); (h) Groud truth.}
\label{compare_cata7}
\end{figure}

To further analyze the performance of the bilinear attention module, we visualize feature maps of its inputs and outputs in Figure~\ref{hotmap}. Compared with input feature maps, output feature maps of the bilinear attention module highlight the region containing the instrument, proving that BAM effectively models semantic dependencies and improves feature representation.
Also, we visualize segmentation results of the network without BAM, which is shown in Figure~\ref{compare_cata7} (e). There is incomplete segmentation in these results, and part of the surgical instrument is identified as background. The network employing BAM achieves excellent results, whose masks are relatively complete and the same as the ground truth.

%

\subsubsection{Ablation Study for Adaptive Receptive Field Module}
Adaptive Receptive Field Module (ARF) is designed to adaptively select the receptive field, adapting to the scale variation of instruments. A range of experiments are set up to verify its excellent performance.

As shown in Table~\ref{ablation}, the network using ARF achieves an increase of 4.97$\%$ mean IOU and 2.94$\%$ mean Dice compared to the basic network. When using BAM, employing ARF brings a 1.50$\%$ increase on mean IOU and 0.87$\%$ increase on mean Dice. Furthermore, ARF only adds 0.1M parameters which only account for 0.46$\%$ of the basic network.
To give a more intuitive result, we visualize some results of the network without ARF, which is shown in Figure~\ref{compare_cata7} (f). The network without ARF has poor segmentation performance than BARNet, indicating the effectiveness of the ARF. The above results suggest that ARF can significantly improve segmentation accuracy with very few parameters.

\begin{figure*}[tbp]
\centering
\subfigure[U-Net]{
\begin{minipage}{0.24\textwidth}
\centering
\includegraphics[width=\textwidth]{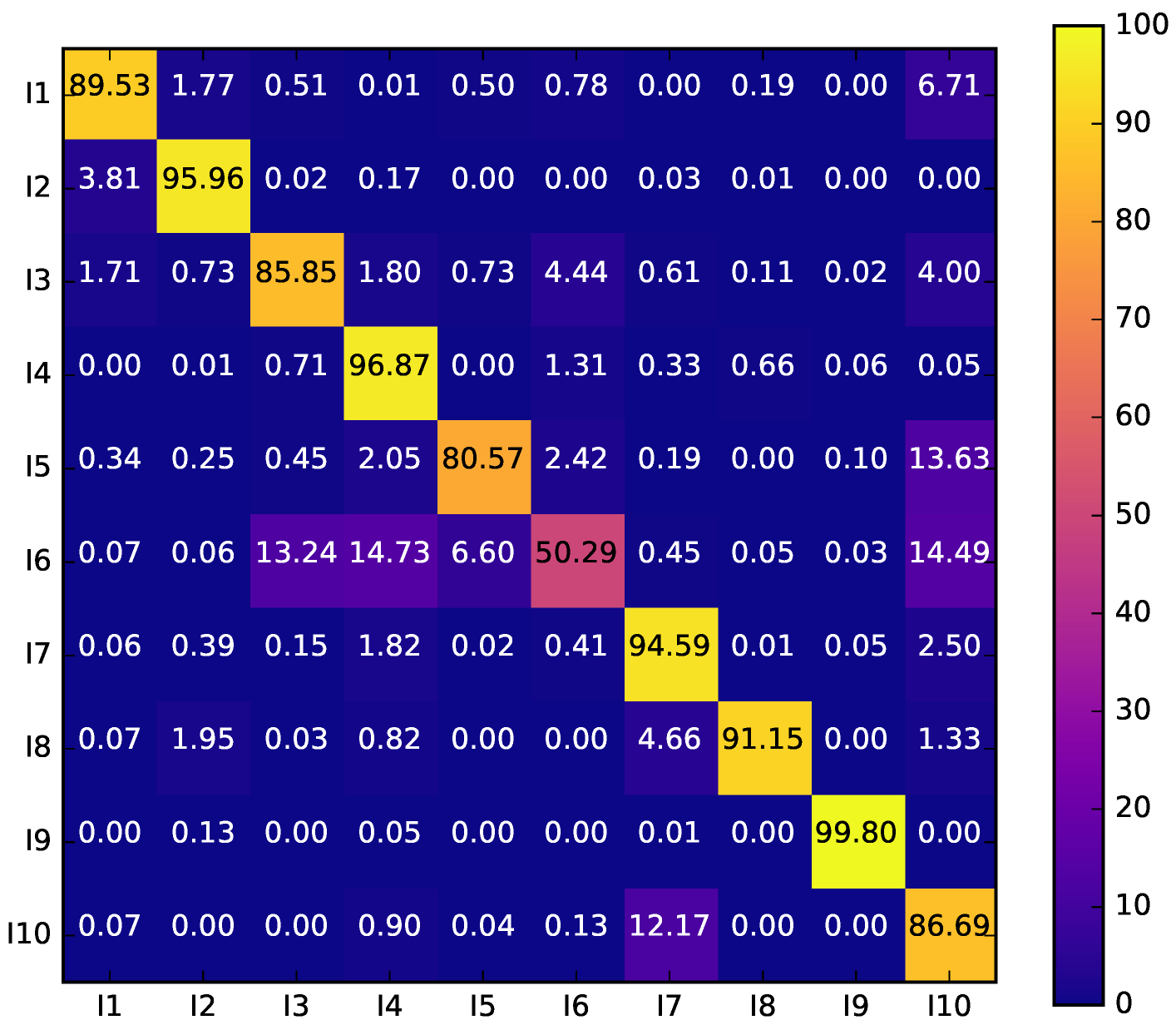}
\end{minipage}
}
\hspace{-0.2cm}
\subfigure[TernausNet]{
\begin{minipage}{0.24\textwidth}
\centering
\includegraphics[width=\textwidth]{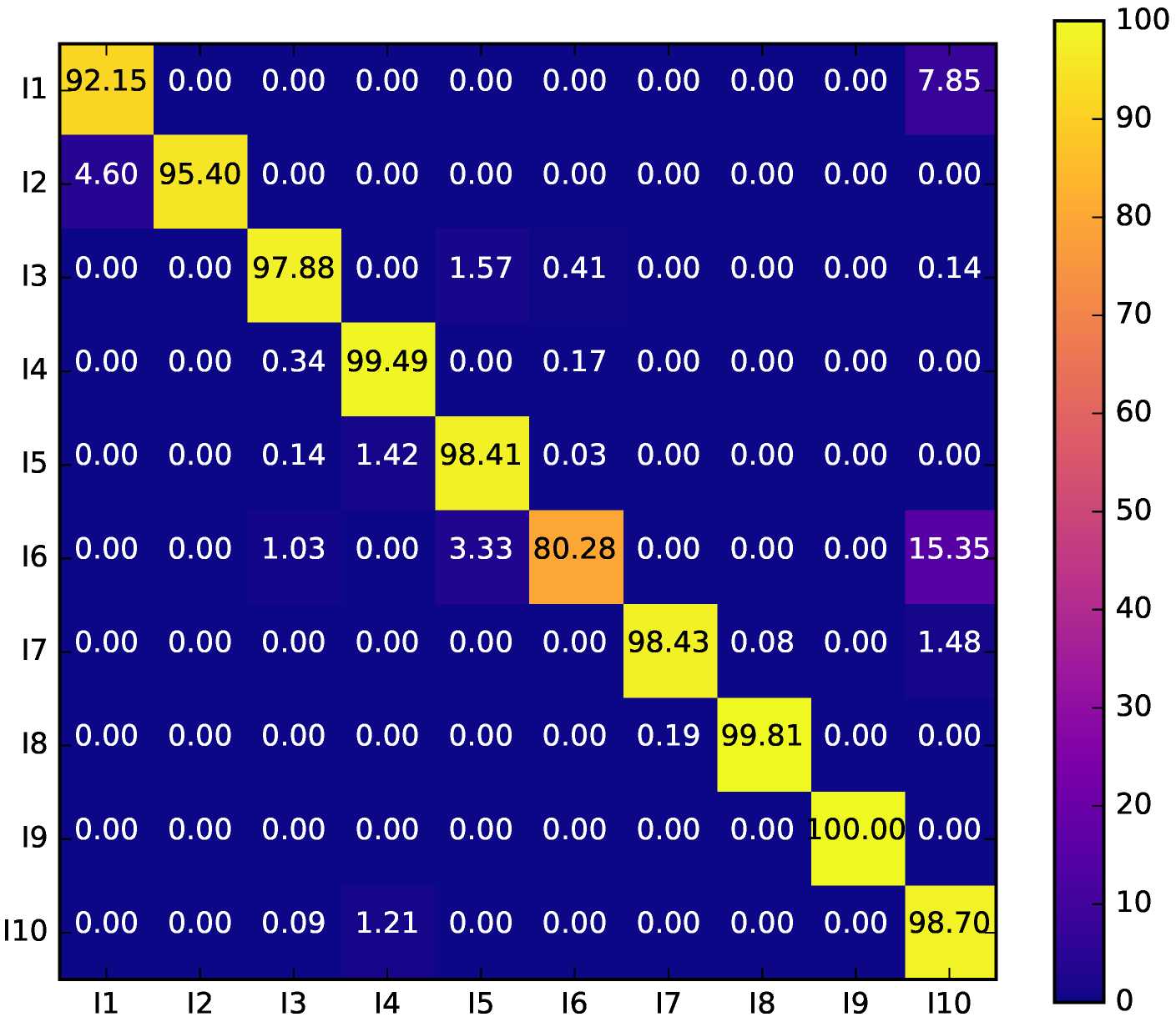}
\end{minipage}
}
\hspace{-0.2cm}
\subfigure[LinkNet]{
\begin{minipage}{0.24\textwidth}
\centering
\includegraphics[width=\textwidth]{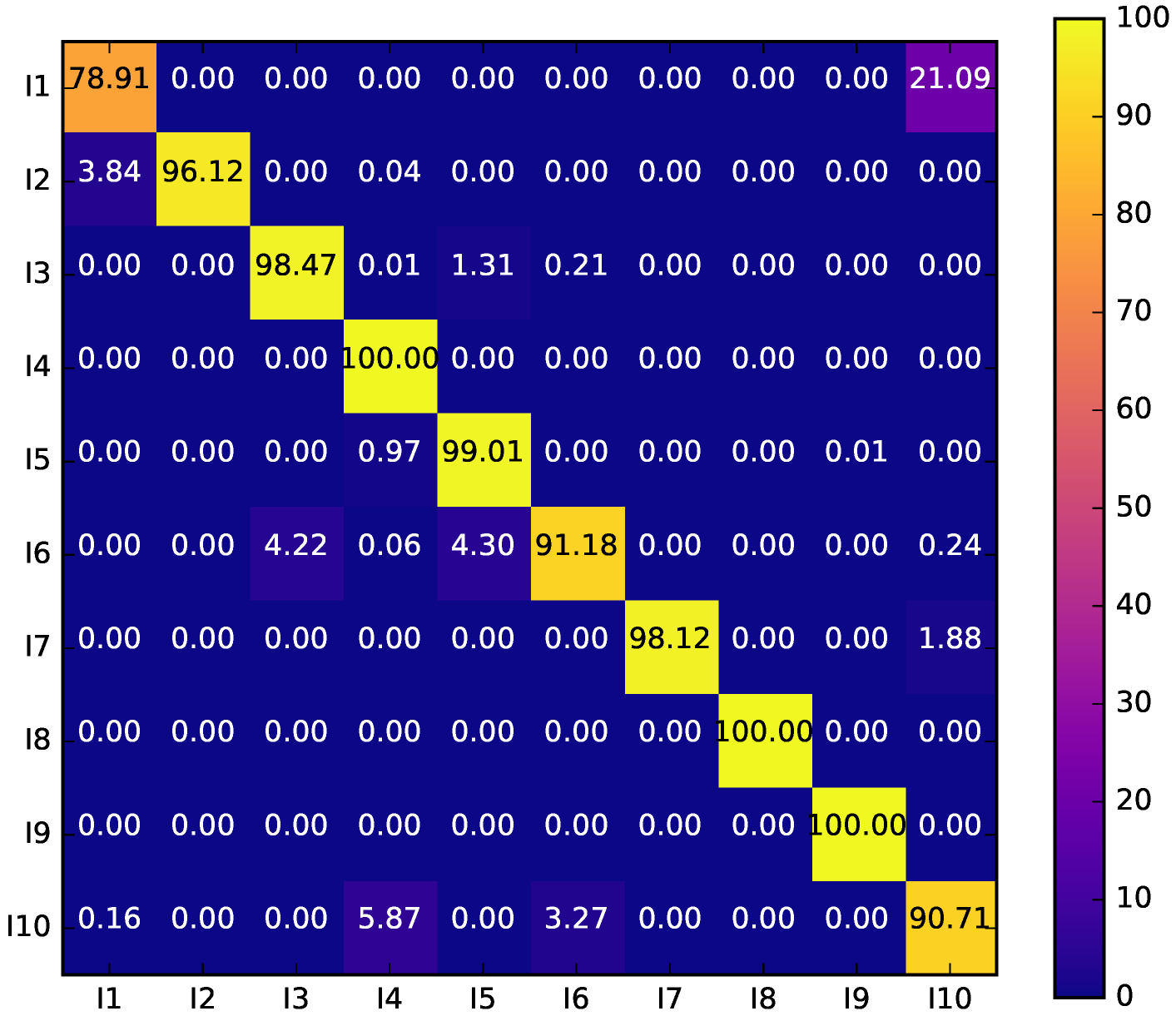}
\end{minipage}
}
\hspace{-0.2cm}
\subfigure[BARNet(Ours)]{
\begin{minipage}{0.24\textwidth}
\centering
\includegraphics[width=\textwidth]{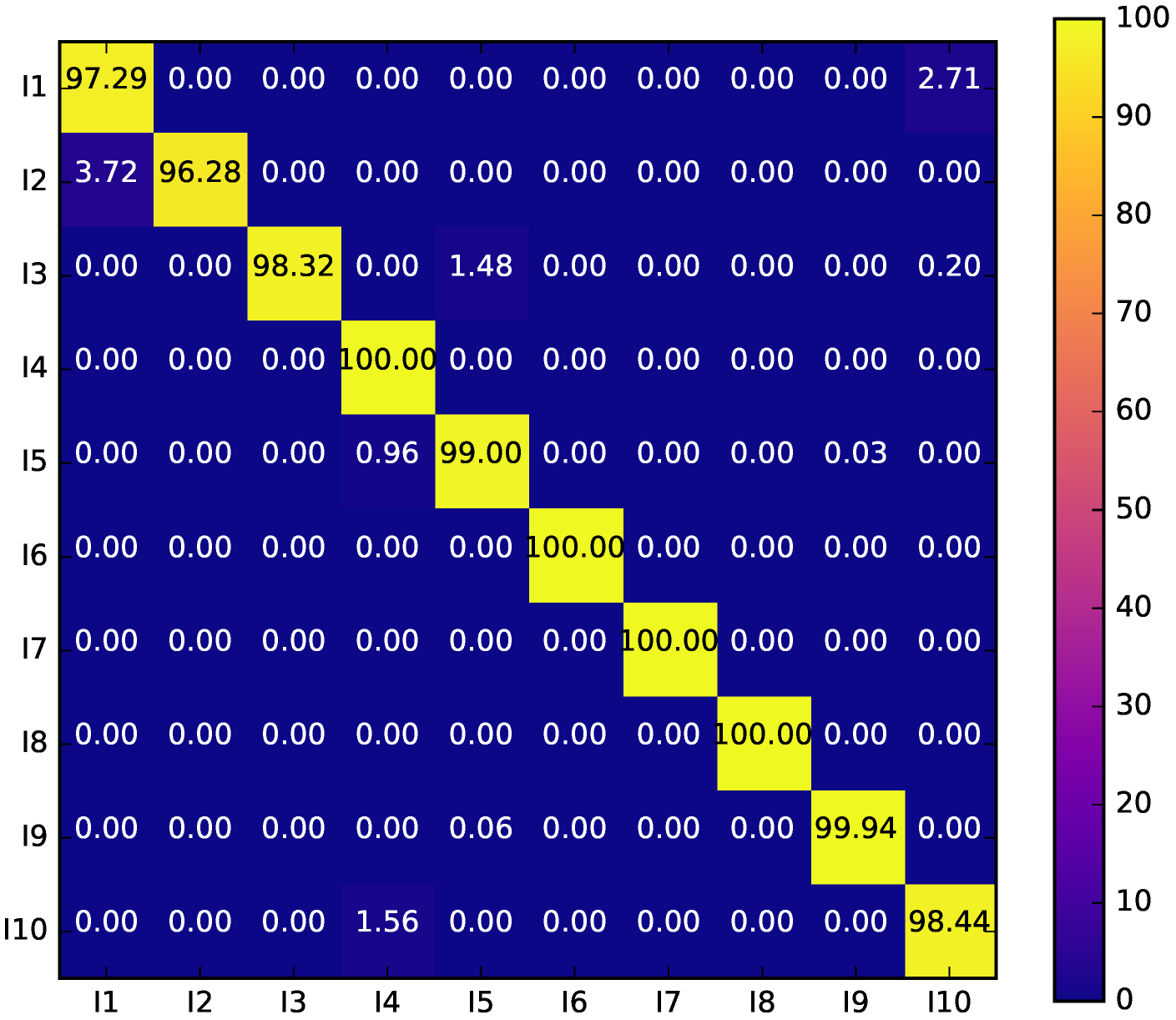}
\end{minipage}
}\setlength{\abovecaptionskip}{0.00cm}
\caption{Pixel accuracy of each class. The X and Y-axis represent prediction and ground truth,
respectively. The numbers on the diagonal are the pixel accuracy of each class. }
\label{confusion}
\end{figure*}


%
\begin{table}[tbp]
  \centering
  \resizebox{\columnwidth}{!}{
    \setlength{\tabcolsep}{0.8mm}{
    \begin{tabular}{c|c|c|c}
    \hline
    \hline
    \textbf{Method}  & \textbf{mDice} & \textbf{mIOU}& \textbf{Param.}\\
    \hline
    U-Net~\cite{unet} &   86.83 & 78.21 & 7.85M \\
    RefineNet~\cite{refinenet}&   93.53   &   88.41    & 25.75M \\
    LinkNet~\cite{linknet} &   94.63 & 91.31 &  21.81M \\
    TernausNet~\cite{ternausnet} &   96.24    &   92.98    & 25.36M \\
    RAUNet~\cite{raunet}&   97.71    &   95.62    & 22.02M \\
    \hline
    BARNet(Ours) &   98.68    &  97.47 &  21.90M \\
    \hline
    \hline
    \end{tabular}%
    }
    }
  \caption{Segmentation results of different methods on Cata7.}
  \label{compare}%
\end{table}%

\begin{table*}[tbp]
  \centering
    \resizebox{\textwidth}{!}{
    \begin{tabular}{c|c|c|c|c|c|c|c|c|c|c|c}
    \hline
    \hline
     & Dataset 1 & Dataset 2 & Dataset 3 & Dataset 4 & Dataset 5 & Dataset 6 & Dataset 7 & Dataset 8 & Dataset 9 & Dataset 10 & mIOU \\
    \hline
    TernausNet & \textbf{0.177} & 0.766 & 0.611 & 0.871 & 0.649 & 0.593 & 0.305 & 0.833 & \textbf{0.357} & 0.609 & 0.542 \\
    ToolNet & 0.073 & 0.481 & 0.496 & 0.204 & 0.301 & 0.246 & 0.071 & 0.109 & 0.272 & 0.583 & 0.337 \\
    SegNet & 0.138 & 0.013 & 0.537 & 0.223 & 0.017 & 0.462 & 0.102 & 0.028 & 0.315 & 0.791 & 0.371 \\
    NCT   & 0.056 & 0.499 & \textbf{0.926} & 0.551 & 0.442 & 0.109 & 0.393 & 0.441 & 0.247 & 0.552 & 0.409 \\
    UB   & 0.111 & 0.722 & 0.864 & 0.68  & 0.443 & 0.371 & 0.416 & 0.384 & 0.106 & 0.709 & 0.453 \\
    UA    & 0.068 & 0.244 & 0.765 & 0.677 & 0.001 & 0.400   & 0.000  & 0.357 & 0.040  & 0.715 & 0.346 \\
    \hline
    Ours& 0.104 & \textbf{0.801} & 0.919 & \textbf{0.934} & \textbf{0.830} & \textbf{0.615} & \textbf{0.534} & \textbf{0.897} & 0.352 & \textbf{0.810} & \textbf{0.643} \\
    \hline
    \hline
    \end{tabular}%
    }
    \caption{Segmentation results on Endovis 2017 dataset. BARNet achieves 64.3$\%$ mean IOU and takes the first place. NCT, UB and UA are the university abbreviation of the participating team. Since the number of samples in each data set is different, each dataset is given a weight related to the number of samples when calculating the mean IOU.}
  \label{endovis2017}%
\end{table*}%

\subsection{Comparison with state-of-the-art on Cata7}
A series of comparative experiments are performed to evaluate the performance of BARNet. BARNet achieves state-of-the-art performance 97.47$\%$ mean IOU and 98.68$\%$ mean Dice, exceeding
the second-ranking method by 1.85$\%$ on mean IOU and 0.97$\%$ on mean Dice. The performance of other methods is much poorer than BARNet, which demonstrates the excellent performance of BARNet.

To further evaluate the segmentation performance of the proposed method for each type of surgical instrument, the confusion matrix about pixel classification is shown in Figure~\ref{confusion}.
We find that our method achieves excellent performance on every type of instrument. Especially, the proposed method outperforms other methods by a significant margin on primary incision knife (I1), lens hook (I6) and bonn forceps (I10). The surface of these surgical instruments is prone to specular reflections due to their special material, making it more difficult to segment them. The bilinear attention module can model complex semantic dependencies and infer the semantic features in specular reflection and shadow regions, addressing the illumination variation issue. Thus, our network achieves better performance on these three instruments.

\subsection{The Results on EndoVis 2017}
To further verify the performance of BARNet, it is evaluated on the Endovis 2017 dataset~\cite{endovis2017}. The test set consists of 10 video sequences. Each sequence contains specific surgical instruments. Datasets 1-8 contain 75 images, respectively. Dataset 9 contains 300 images, and the number of images is the same in dataset 10. The test results are reported in Table~\ref{endovis2017}. TernausNet~\cite{ternausnet}, ToolNet~\cite{toolnet} and SegNet~\cite{segnet} are evaluated on EndoVis 2017. The test results of other methods are from the MICCAI EndoVis challenge 2017~\cite{endovis2017}.

BARNet achieves 64.30$\%$ mean IOU, which outperforms existing methods. The second-ranking method, TernausNet, achieves 54.20$\%$ mean IOU. Compare with this method, our network achieves 10.10$\%$ gain on mean IOU. The performance of other methods is much poorer than BARNet. Furthermore, BARNet achieves the best results in 7 video sequences and takes second place in the other three video sequences. These results show that BARNet achieves state-of-the-art performance on this dataset.

\begin{figure}[tbp]
\centering
\includegraphics[width=0.98\columnwidth]{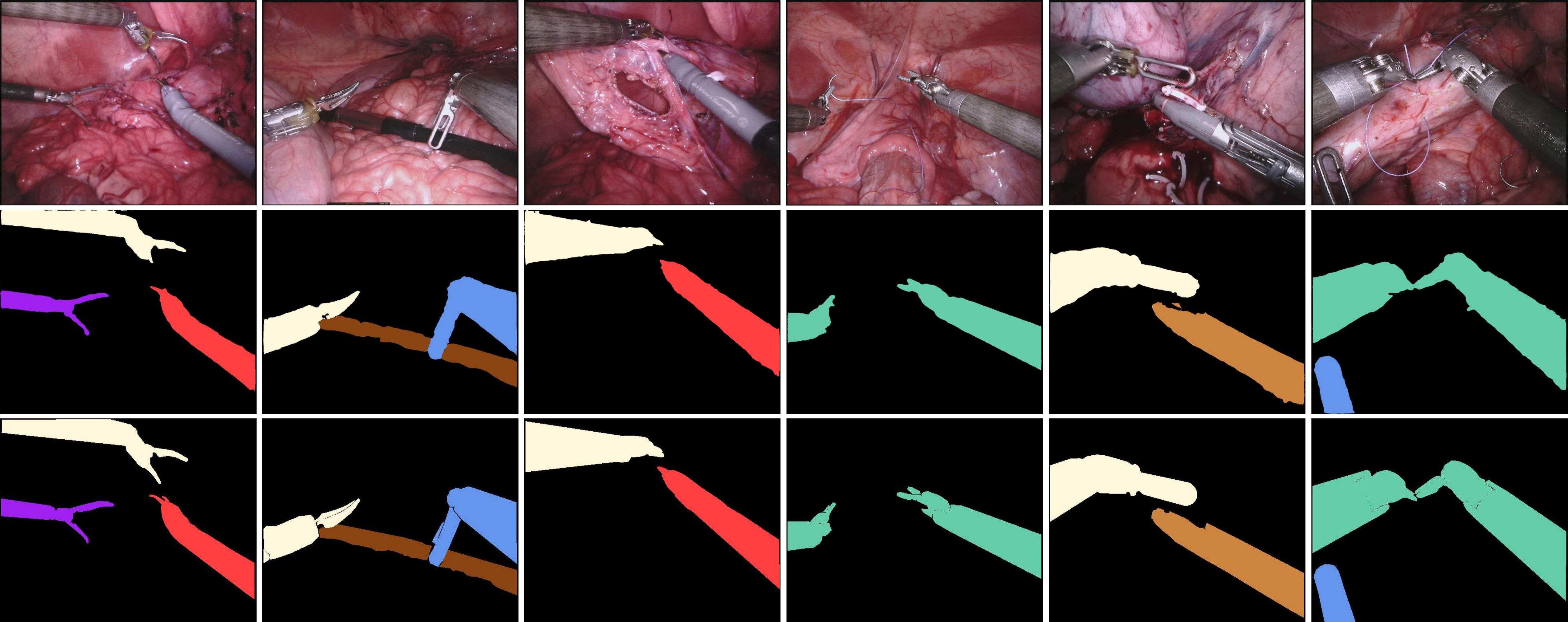}
\caption{Visualization for segmentation results of BARNet. From top to bottom: original images, segmentation results, ground truth.}
\label{endovis_result}
\end{figure}

To give intuitive results, the segmentation results of BARNet are visualized in Figure~\ref{endovis_result}. We find multiple specular reflections and shadow areas in the figure. Besides, the scale and shape of surgical instruments are significantly varied. Despite these challenges, BARNet can still accurately segment surgical instruments, whose segmentation results are basically consistent with the ground truth.
\section{Conclusion}
In this paper, the bilinear attention network with adaptive receptive field (BARNet) is proposed for surgical instrument segmentation. Bilinear attention module captures global contexts and second-order statistics to improve feature representation. Adaptive receptive field selects feature maps with specific sizes to choose appropriate receptive fields. A series of ablation experiments prove that the bilinear attention module and adaptive receptive field module contribute to improving network performance. Moreover, BARNet achieves state-of-the-art performance on both Cata7 and EndoVis 2017.

\bibliographystyle{named}
\bibliography{ijcai20}

\end{document}